\documentclass[pdflatex,sn-mathphys]{sn-jnl}

\usepackage{placeins}
\usepackage{caption}
\usepackage{subcaption}

\jyear{2022}%

\theoremstyle{thmstyleone}%
\theoremstyle{thmstyletwo}%

\theoremstyle{thmstylethree}%

\begin{document}

\title[Gender bias in blockbuster movies]{Identifying gender bias in blockbuster movies through the lens of machine learning.}


\author*[1]{\fnm{Muhammad Junaid} \sur{Haris}}\email{Junaid.Haris@abo.fi}

\author[1]{\fnm{Aanchal} \sur{Upreti}}\email{Aanchal.Upreti@abo.fi}

\author[1]{\fnm{Melih} \sur{Kurtaran}}\email{Melih.Kurtaran@abo.fi}

\author[2]{\fnm{Filip} \sur{Ginter}}\email{Filip.Ginter@utu.fi}

\author[1]{\fnm{Sebastien} \sur{Lafond}}\email{Sebastien.Lafond@abo.fi}

\author[1]{\fnm{Sepinoud} \sur{Azimi}}\email{Sepinoud.Azimi@abo.fi}

\affil[1]{\orgdiv{Faculty of Science and Engineering}, \orgname{Åbo Akademi Univeristy}, \country{Finland}}

\affil[2]{\orgdiv{TurkuNLP}, \orgname{University of Turku}, \country{Finland}}

\abstract{The problem of gender bias is highly prevalent and well known. In this paper, we have analysed the portrayal of gender roles in English movies, a medium that effectively influences society in shaping people's beliefs and opinions. First, we gathered scripts of films from different genres and derived sentiments and emotions using natural language processing techniques. Afterwards, we converted the scripts into embeddings, i.e. a way of representing text in the form of vectors. With a thorough investigation, we found specific patterns in male and female characters' personality traits in movies that align with societal stereotypes. Furthermore, we used mathematical and machine learning techniques and found some biases wherein men are shown to be more dominant and envious than women, whereas women have more joyful roles in movies. In our work, we introduce, to the best of our knowledge, a novel technique to convert dialogues into an array of emotions by combining it with Plutchik's wheel of emotions. Our study aims to encourage reflections on gender equality in the domain of film and facilitate other researchers in analysing movies automatically instead of using manual approaches.}

\keywords{Gender bias, Machine learning, Natural Language Processing, Gender stereotypes.}



\maketitle

\section{Introduction}\label{sec1}

In our work, we aim to study and analyse gender bias based on emotions expressed by male and female characters in movies using different Natural Language Processing (NLP) techniques. For this purpose, we chose around thirty blockbuster English movies from IMDb  and performed a comparative analysis of the characters. The study is mainly based on the statistical distribution of male and female characters over a certain period and the sentiment and emotions expressed in their dialogues. Through our work, we are trying to present an approach to understanding and promoting studies on the highly prevalent issue of gender inequality in movies. Furthermore, our analysis allows us to help address the global problem of gender inequality and bring positive social change using NLP.

\section{Related Works}\label{sec2}
Several studies have emerged in the past two decades that attempt to assess the gender gap between men and women in various fields \citep{lariviere_bibliometrics_2013, noauthor_boxed_nodate, wagner_its_2015}.

Many studies have utilised the Bechdel Test \cite{bechdel2010dykes} to evaluate gender bias in movies. Kagan et al. \citep{kagan_using_2020} in their studies presented 15,540 movie social networks where their findings showed a gender gap in almost all genres of the film industry. However, they also found that all aspects of women’s roles in movies show a trend of improvement over the years with a constant rise in the centrality of female characters and the number of movies that pass the Bechdel test.

Xu et al. \citep{xu_cinderella_2019} in their analysis of 7,226 books, 6,087 film synopsis, and 1,109 film scripts, identified the constructed emotional dependence of female characters on male characters, also called the Cinderella complex, where women depend on men for a happy and fulfilling life. They used the word embedding techniques with word vectors trained on the Google News dataset to construct a happiness vector for automatically calculating the happiness score for every word in the document. The happiness vector contains words such as \emph {success, succeed, luck, fortune, happy, glad, joy, and smile} for positive and \emph {failure, fail, unfortunate, unhappy, sad, sorrow, and tear} for negative sentiment. Their analysis of female and male characters showed that female's word vector is oriented toward romance, whereas men's toward adventure. Their studies showed how such narratives embed stereotypical gender roles into moral tales and institutionalise gender inequality through these cultural products.

Similarly, Yu et al. \citep{yu_emotion_2017} analysed the emotion represented in Korean thriller movie scripts. The authors performed manual emotion annotation on movie scenes based on eight emotion types defined by Plutchik \citep{plutchik1988nature}. Further, they used the Python-based NLTK VADERSentiment tool to analyse the sentiment of the same script and compared the results obtained with manual tagging. Their result showed that the emotions of anger and fear were most matched, whereas the emotion of surprise, anticipation, and disgust had a lower matching score.

Anikina \citep{anikina_sentiment_nodate} in her work performed manual annotation on movie dialogues based on Plutchik's emotion wheel. The author used machine learning classifiers (fastText, OpenNLP and SVM) to train on different datasets and implemented rule-based NRC emotion lexicon classifiers to detect the emotions. The study compared the results obtained from NRC Lexicon with manually annotated data to find the accuracy. The study highlighted how the unavailability of annotated resources on movie script domain affected the training process as the data from different domains had to be used and also discussed the limitations of used classifiers in performing multi-label emotion classification.

Similarly, the Bechdel test mostly focuses on assessing the fairness of female representation by three rules which fails even for female oriented movies \citep{noauthor_why_nodate, noauthor_25_nodate, noauthor_media-research_nodate} and also does not address stereotypes. In our work, we address the stereotypes through sentiment and emotion analysis, which gives us the benefit of tackling the limitations of the Bechdel test. Moreover, in previous studies, stereotypes are generally only analysed using word embedding techniques by creating word vectors. In our work, we are not only relying on the word vectors but also analysing the sentiment and emotions embedded with those words for better representation of our results and hence expanding the work beyond just the positive-negative affect dimension. Further, we present a novel approach to converting dialogues into an array of emotions by combining it with Plutchik's wheel of emotions.

\section{Methods and Experiment}\label{sec3}
\begin{figure}[H]
    \centering
    \includegraphics[width=7cm]{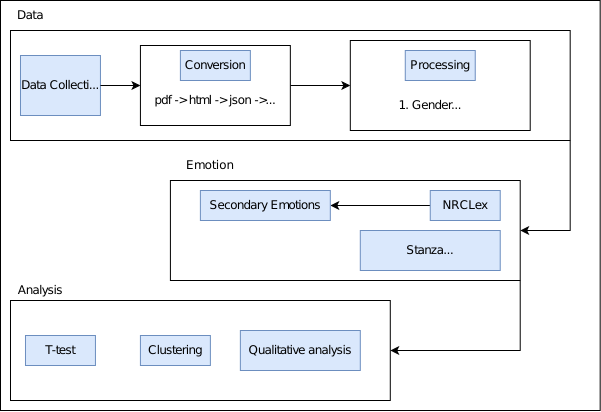}
    \caption{Workflow of the study including three modules: (1) data processing module, (2) emotion detection module, and (3) analysis module.}
    \label{fig:flow_chart}
\end{figure}

In this Section, we present the workflow of our approaches, as presented in Figure \ref{fig:flow_chart}. A brief description of the three modules is as follows. More detailed information on each module is provided in Sections~\ref{sec:data_processing_module}, \ref{sec:emotion_recognition_module} and \ref{sec:analysis_module} respectively.

\paragraph{Data Processing Module}
In this module, we convert the scripts of movies from PDF into a machine-readable format. Movie scripts are first converted from PDF into HTML using online tools. Then we process the HTML and separate scenarios and dialogues of each character.

\paragraph{Emotion Recognition Module}
This module looks at every character's prevalent positive and negative sentiments using Stanza \citep{qi2020stanza}, a Python natural language analysis package. Next, we run NRCLex \citep{Mohammad13} -- a rule-based model for emotion detection trained on 27,000 manually annotated common words and phrases. From this step, we get the "emotion scores" for eight primary emotions: fear, anger, trust, surprise, sadness, disgust, joy, and anticipation. Finally, with the help of Plutchik's wheel of emotions, which describes how emotions are related, we compute 24 secondary emotions. In this module, we convert sentences into embeddings. An embedding is a learned numerical representation of sentences in the form of a vector, where similar sentences will have a similar representation.

\paragraph{Analysis Module}
In the last module, we perform empirical analysis, statistical methods such as Mann-Whitney U test, and machine learning techniques, namely clustering and classification, to determine if gender bias exists in the collection of scenarios. Our approach is modular, and modules are loosely coupled and highly cohesive. 

\subsection{Data Processing Module}~\label{sec:data_processing_module}
In this Section, we present the first module of this study, i.e., the data processing module. This module includes data collection and data processing stpng.
\paragraph{Data Collection}~\label{sec:data_collection}

\begin{figure}[H]
    \centering
    \includegraphics[width=7cm]{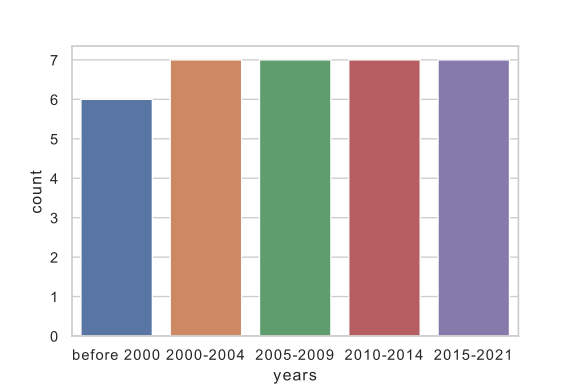}
    \caption{Year Distribution of the Movies Selected.}
    \label{fig:movie_year}
\end{figure}

To ensure the global reach of the movies and their impact on society, we opted only for the top-rated movies of all time listed on IMDB. We downloaded scripts of thirty-four movies from different genres, namely romance, fantasy, fiction, drama, and action, spread over the years ranging from 1972 to 2021. While choosing the scripts, we considered attributes such as accessibility, the format of the script, and its impact. The year distribution of the movies is presented in Figure \ref{fig:movie_year}. In total, we collected 26279 dialogues and 457 characters, out of which 118 are female characters and 339 are male.

\paragraph{Data Processing}~\label{data_processing}

\begin{figure}[H]
    \centering
    \includegraphics[width=6cm]{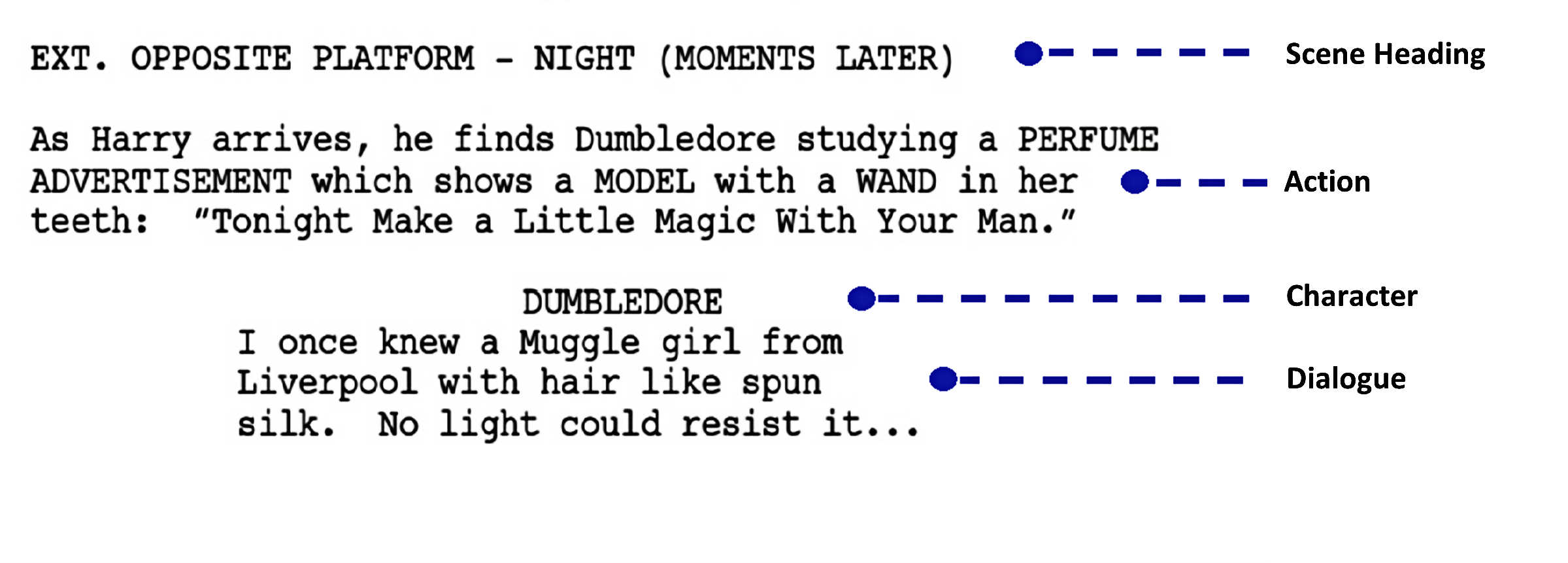}
    \caption{How movie scripts are typically written}
    \label{fig:movie_script}
\end{figure}

Movie scripts are technical documents created by screenwriters and serve as an essential reference for filmmakers. Blocks of text divided into scene headers, scene descriptions, speakers, and utterances are parts of a standard movie script. The critical point is the different indentations used for every block. The typical format of scripts is shown in Figure \ref{fig:movie_script}.

We converted these scripts into a processed set of scenarios and dialogues where individual characters' information is extracted and exploited using the information about indentation. We would refer to them as character dictionaries for the rest of this paper. We first converted PDF files into HTML format and used BeautifulSoup  --  a Python library for pulling data out of HTML -- and regex to create character dictionaries. In the HTML files, the style attribute of every element has a property named `left' and `top', which indicates the value of the starting pixel from the left, i.e. indentation and dialogue sequence, respectively. We have stored the processed version of scripts in JSON format, i.e. attribute-value pairs, where the attribute is the name of the character and the value is the list of dialogues.




In the next step, we created a database from the character dictionaries by further processing the data. Screenwriters sometimes add extra text to the character's name, such as (V.O.), (O.S.), etc., suggesting a voice-over or off-screen dialogue. First, we processed such terms and deleted this extra information because it is irrelevant to our study. Then, we tagged each character with their respective genders.




Furthermore, we added the movie's name and the year it was released into the database for visualisation and analysis purposes. Finally, we dropped the characters with less than five dialogues since they did not contribute to our results in a meaningful way. Performing all these stpng, we generated a final database, which was also manually validated to ensure its correctness.
\subsection{Emotion Recognition Module}~\label{sec:emotion_recognition_module}
In this Section, we present the second module of this study, i.e., the emotion recognition module. This module includes finding sentiments and emotions for each character.

\subsubsection{Sentiment Analysis}~\label{sentiment_analysis}

We first used the Stanza package \citep{qi2020stanza} to find the sentiment of every dialogue. Stanza is a Python natural language analysis package that supports sentiment analysis on raw text from diverse sources as input. It produces annotations without having to annotate or tokenize the text manually. It is built with a highly accurate CNN classifier pipeline trained on 112 datasets, enabling efficient training and evaluation of annotated data. The reason for using Stanza is that it is built on top of the PyTorch library \citep{noauthor_pytorch_nodate}, which gives a high-speed performance on GPU-enabled machines and has better accuracy than other under-optimized models. Stanza classified each dialogue as positive, negative, or neutral. However, more than 70\% of the conversations organised by Stanza were neutral. Therefore, Stanza did not provide much information about the bias in male or female characters.


These results prompted us to check the validity of the stanza model, for which we used the trivial approach of manually annotating the data ourselves and then comparing the result with the stanza model. To reduce the bias, the same data were manually annotated by two different people, and 179 dialogues spoken by Harry in the Harry Potter series were chosen randomly. After comparing the accuracy of the model with human evaluation, we acquire the following accuracy scores shown in Table
\ref{tab:sentiment_verification}.
\begin{table}[h]
\begin{center}
\begin{minipage}{174pt}
\caption{Accuracy of Stanza}\label{tab:sentiment_verification}
\begin{tabular}{@{}llll@{}}
\toprule
Accuracy & Stanza  & Person 1 & Person 2\\
\midrule
Stanza & 1.0 & 0.71  & 0.78  \\
Person 1 & 0.71 & 1.0 & 0.74  \\
Person 2 & 0.78 & 0.74  & 1.0  \\
\botrule
\end{tabular}
\footnotetext{Note: This is the accuracy after manually annotating the data.}
\end{minipage}
\end{center}

\end{table}

\subsubsection{Emotion Detection}~\label{emotion_detection}

\begin{figure}[H]
    \centering
    \includegraphics[width=7cm]{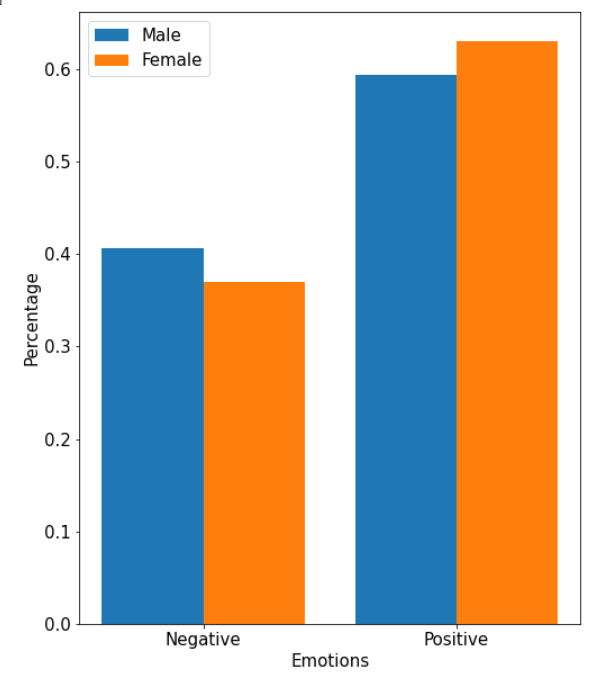}
    \caption{Positive and Negative Emotion Scores}
    \label{fig:emotion_pn}
\end{figure}

By looking at the results of Stanza, we came to know that the overall outlook of positivity and negativity was not drastically different. But, generally, the emotions conveyed by humans are much more complex than only positive and negative. Therefore, for deeper analysis, we ran NRCLex on every dialogue. NRCLex takes a string, i.e. dialogue, as input and returns the emotion scores for the primary eight emotions. Those primary emotions can also be grouped into positive and negative, as suggested by Robert Plutchik. The model classifies joy, anticipation, trust, and surprise as positive emotions and anger, fear, sadness, and disgust as negative emotions. Comparison of genders based on positive vs negative emotions represented in Figure \ref{fig:emotion_pn}.

By running NRCLex, we converted every dialogue into an embedding of eight dimensions. Each dimension represents one primary emotion, and its value is a floating-point number showing the likelihood of that sentence conveying that emotion. It's a probability distribution. Therefore, the embedding or vector sums up to one.

\begin{figure}[H]
    \centering
    \includegraphics[width=7cm]{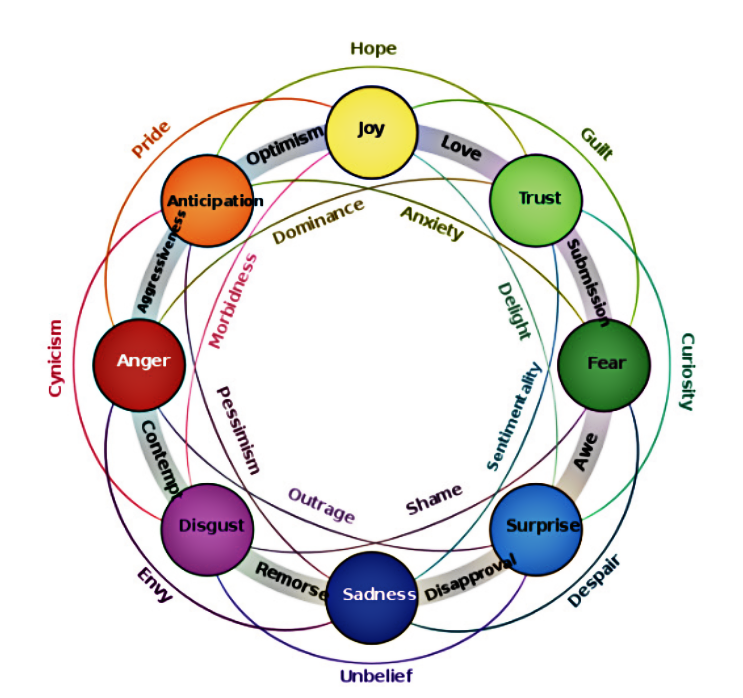}
    \caption{Plutchik's wheel of emotion}
    \label{fig:wheel_of_emotion}
    \citep{Mohsin2019/05}
\end{figure}

Furthermore, we analysed these primary emotions based on Plutchik's emotion wheel as shown in Figure \ref{fig:wheel_of_emotion}. Psychologist Robert Plutchik proposes it through his famous ``wheel of emotions'' \citep{Mohsin2019/05}. The model shows the intensities of different emotions and how they are interconnected. We use this model as a reference to create complex emotions.

We computed 24 secondary emotions by averaging the scores of two primary emotions, e.g. envy was derived by taking an average score of sadness and anger. Lastly, we augmented this new information with the previous database and converted every dialogue into a vector of dimension 32. The emotions are; anger, joy, anticipation, surprise, trust, delight, sadness, disgust, hope, curiosity, despair, confined, envy, cynicism, pride, love, submission, shame, awe, disapproval, remorse, aggression, anxiety, outrage, fear, dominance, guilt, sentimentality, optimism, pessimism, contempt, and morbidness.




\subsection{Analysis Module}~\label{sec:analysis_module}
In this Section, we present the third module of this study, i.e., the analysis module. This module describes major trends in the data and how characters are written.

\subsubsection{Clustering}~\label{clustering}

Clustering is the process of splitting a population or set of data points into many groups so that data points in the same group are more similar than data points in other groups. To put it another way, the goal is to separate groups with similar characteristics and assign them to clusters.

In our final database, we averaged the emotion scores of characters based on all of their dialogues. This new database has a size of 457x32, where 457 is the total number of characters. In our database, the ratio of male and female characters is roughly 3:1, i.e., for every female character, there are three male characters. The idea behind applying clustering is that if there is no bias in how male and female characters are written, the clusters made should also follow more or less the same ratio as there is in the data.

We use both k-means clustering and hierarchical clustering. First of all, we used the elbow method on the sum of squared distances to find the optimal number of clusters, which turned out to be seven. For Agglomerative clustering, we use euclidean distance and ward's linkage, a method for hierarchical clustering. The results of clustering are discussed in Section 4.5.


\section{Results}
This Section discusses the findings of our analysis using Empirical, Machine Learning, and statistical techniques.

\subsection{Gender Distribution over the years}
\begin{figure}[H]
    \centering
    \includegraphics[width=7cm, height=7cm]{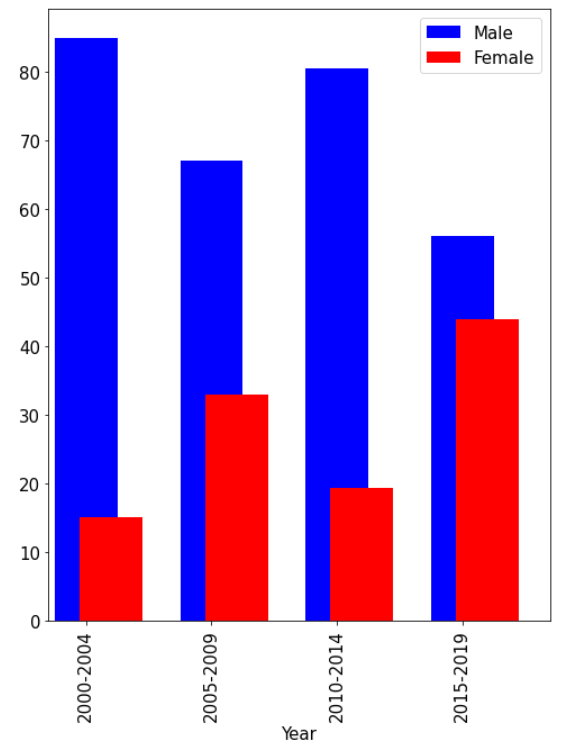}
    \caption{Gender Distribution based on their presence in movies}
    \label{fig:gender_overtime}
\end{figure}

We have compared the ratio of male and female distribution over the years based on the number of their presence and based on the number of their dialogues. Figure \ref{fig:gender_overtime} shows grouped total number of characters over the last 20 years. During the years 2000-2004, only 15.1\% of characters were females however, during 2015-2019, this percentage increased to 43.9\%. It is observed that with the evolution of time, gender distribution is getting more or less balanced. 

\subsection{Mann-Whitney U test}
Based on NRCLex, we have extracted the emotions expressed by male and female characters from their dialogues in all the movies. The obtained emotion score was within the range of 0-1. For detailed analysis, we performed the Mann-Whitney U test -- a form of inferential statistic used to see if there is a significant difference between two groups' means -- on all 32 emotions in the two gender groups. We choose the Mann-Whitney U test because it is non-parametric and robust against non-normality, and our data is not normal. Through this comparison, we found some notable differences in the distribution of some emotions. The analysis showed that male and female characters exhibit most emotions like anger, aggressiveness, despair, envy, outrage, and love very differently from one another. Male characters have higher values for emotions like anger and aggressiveness.
On the other hand, female characters have higher values for emotions such as joy. These results are in line with the gender stereotypes in our society. Women are perceived to be more loving and caring, whereas men are supposed to be more aggressive and powerful. 

The U1, U2, and P-value scores of some emotions are shown in Table \ref{tab:T-test}.

\begin{table}[h]
\begin{center}
\begin{minipage}{174pt}
\caption{U1, U2, and P-value score on various emotions}\label{tab:T-test}
\begin{tabular}{@{}llll@{}}
\toprule
Emotions & U1 Score & U2 Score  & P-values \\
\midrule
			
Anger & 68795197.5 & 64719982.5 & 2.935138e-11\\ 
Joy & 65605789.0 & 67909391.0 &  5.811094e-04 \\ 
Aggressiveness &  67567442.5 & 65947737.5 & 2.757313e-05 \\
Outrage &  67366511.5 & 66148668.5 & 3.886770e-04 \\ 

\botrule
\end{tabular}
\end{minipage}
\end{center}
\end{table}

\begin{figure}
\begin{subfigure}{.5\textwidth}
  \centering
  \includegraphics[width=.8\linewidth]{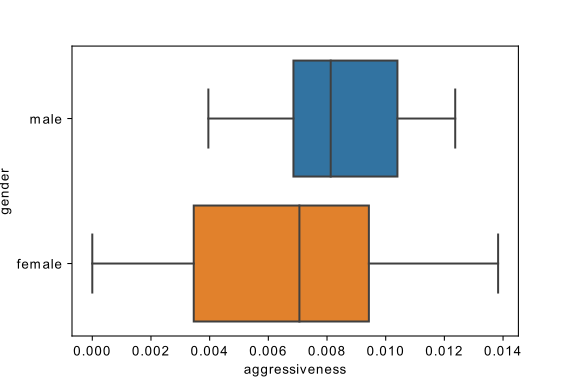}  
  \caption{Aggressiveness}
  \label{fig:sub-first}
\end{subfigure}
\begin{subfigure}{.5\textwidth}
  \centering
  \includegraphics[width=.8\linewidth]{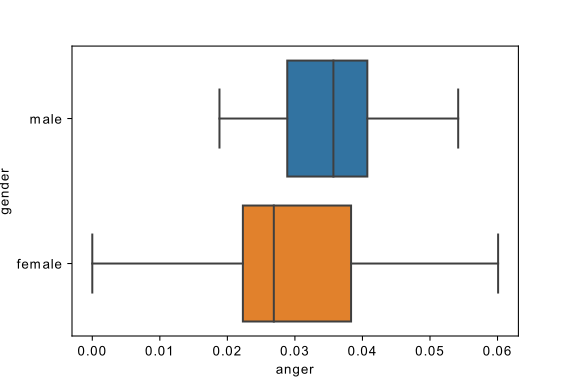}  
  \caption{Anger}
  \label{fig:sub-second}
\end{subfigure}
\newline
\begin{subfigure}{.5\textwidth}
  \centering
  \includegraphics[width=.8\linewidth]{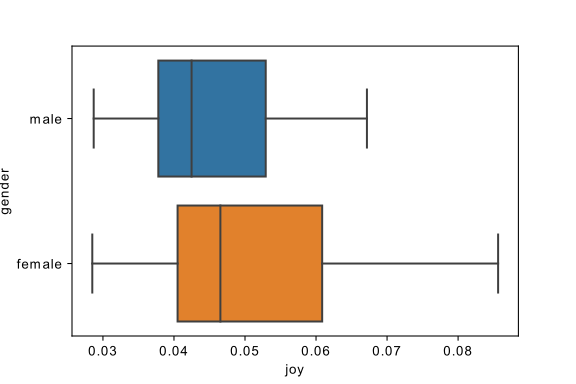}  
  \caption{Anxiety}
  \label{fig:sub-third}
\end{subfigure}
\begin{subfigure}{.5\textwidth}
  \centering
  \includegraphics[width=.8\linewidth]{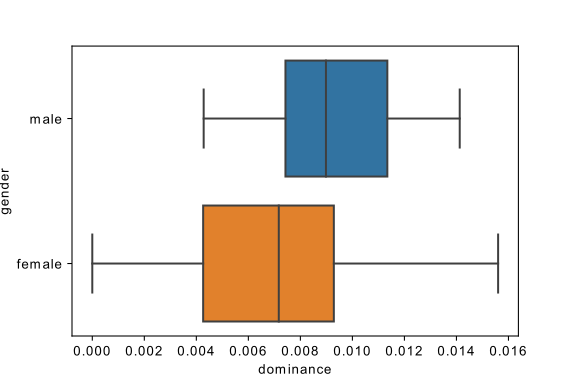}  
  \caption{Dominance}
  \label{fig:sub-fourth}
\end{subfigure}
\caption{Box Plots on various emotions}
\label{fig:boxp}
\end{figure}

\subsection{Visualising Data}

Box plots comparison of female vs male on average emotion scores is shown in Figure \ref{fig:boxp}. Female plots are shown in blue colour, and the male plot is shown in red colour. According to our results, it is observed that male characters have higher scores on the emotions such as aggressiveness and dominance, and female characters have higher scores on emotions such as Joy. 

\begin{figure}[H]
    \centering
    \includegraphics[width=7.5cm]{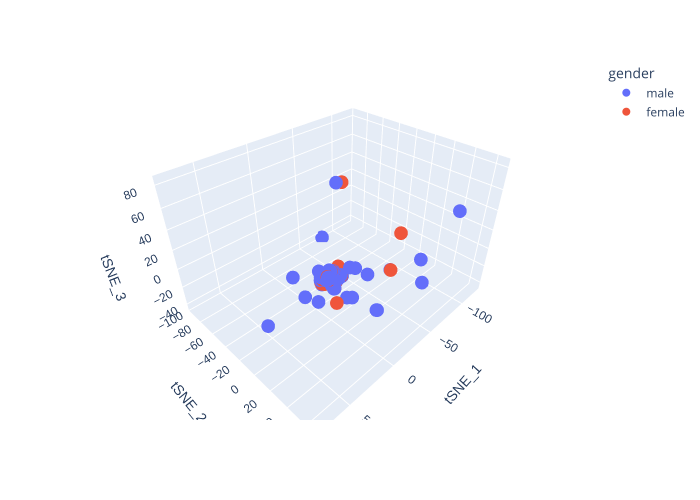}
    \caption{Plotting characters based on emotions using t-SNE}
    \label{fig:clus}
\end{figure}

We have used the t-SNE dimensionality reduction technique to visualise the characters' data. It calculates a similarity measure based on the distance between points and maintains the global structure. In other words, it converts some N-dimensional data into k-dimensional data, where $k<N$ and $x_1, x_2, \ldots, x_K$ are the new axes for the data. In the visualisation shown in Figure \ref{fig:clus} male characters are represented as blue and female characters as red. It is observed that most females are crowded at the centre, whereas the males are clustered sparsely. This uneven distribution shows a pattern in which characters are primarily written in movies. With the exception of two, most female characters exhibit more or less the same emotions, implying they are written in the same way. In contrast, male characters in the film exhibit a range of emotions and have much more diverse personality traits showing a difference in how they are written.

\subsection{Dialogues Analysis}
In order to analyse the variations in the dialogues spoken by male and female characters, we performed a sentence-level analysis which involved extracting the most commonly used words by both genders.

\begin{figure}
\begin{subfigure}{.5\textwidth}
  \centering
  \includegraphics[width=.8\linewidth]{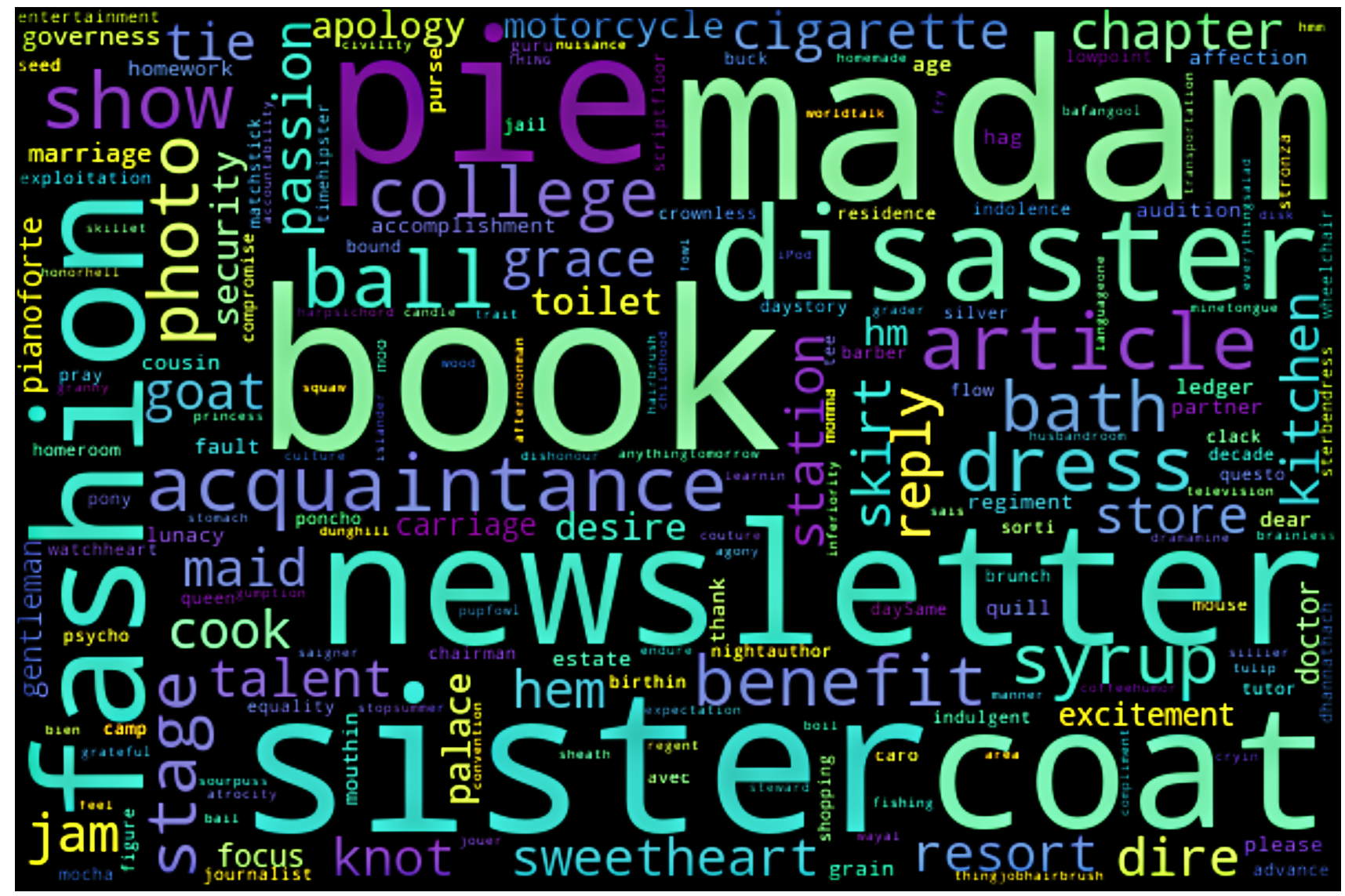}  
  \caption{Female}
  \label{fig:sub-first}
\end{subfigure}
\begin{subfigure}{.5\textwidth}
  \centering
  \includegraphics[width=.8\linewidth]{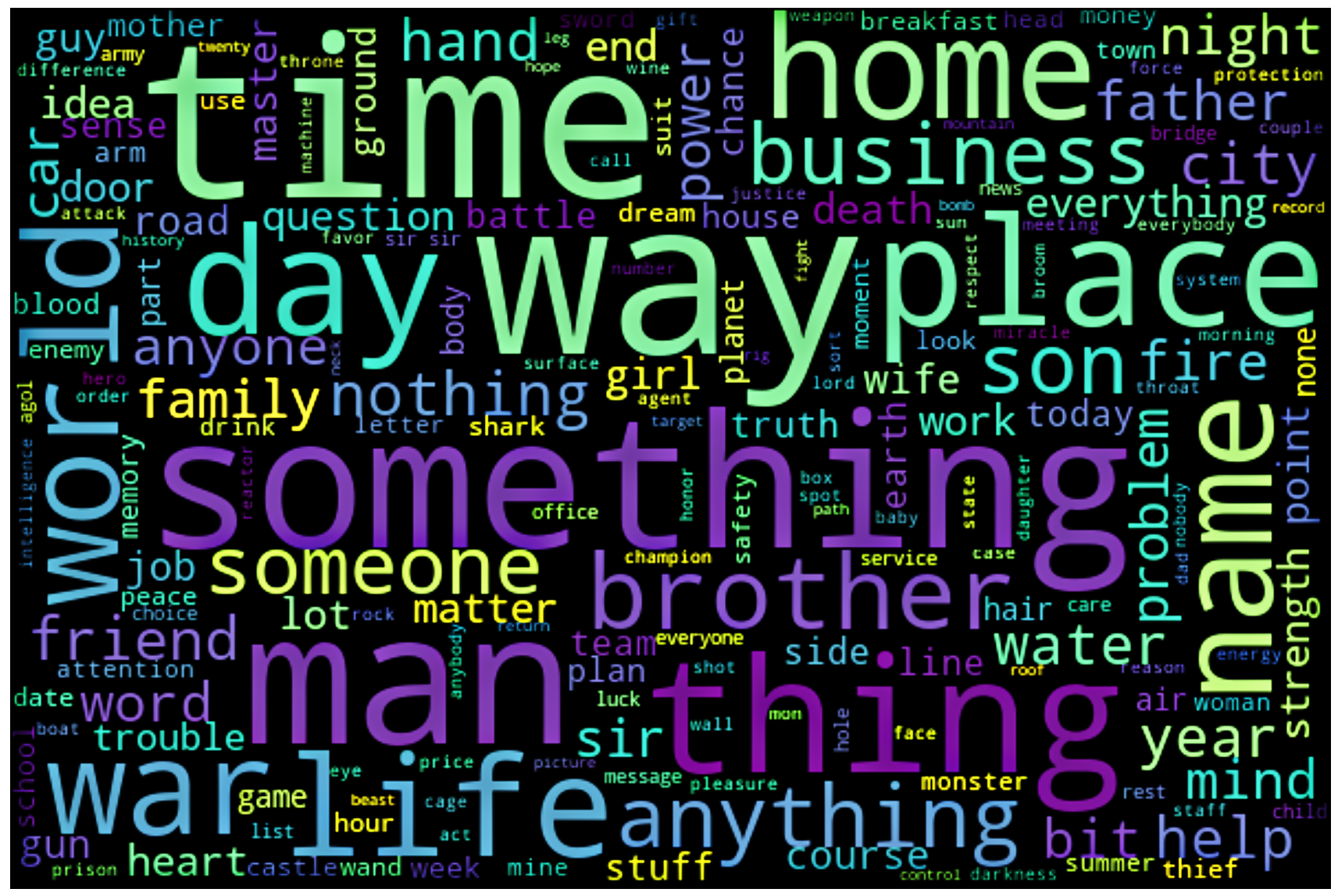}  
  \caption{Male}
  \label{fig:sub-second}
\end{subfigure}
\caption{Distribution of non-overlapping nouns in the dialogues of female and male characters}
\label{fig:wordcloud}
\end{figure}

In Figure \ref{fig:wordcloud}, the word clouds, matching nouns used by male and female characters were excluded in order to trace the unique differences between each gender. In this illustration, we can notice that the female characters' dialogues commonly included the nouns: kitchen, fashion, dress, skirt, sweetheart, and madam; meanwhile, the male characters used the nouns: time, business, war, world, man, and home. These results correlate with gender stereotypes in our society.

\subsection{Clustering}
\begin{figure}
\begin{subfigure}{.5\textwidth}
  \centering
  \includegraphics[width=.95\linewidth]{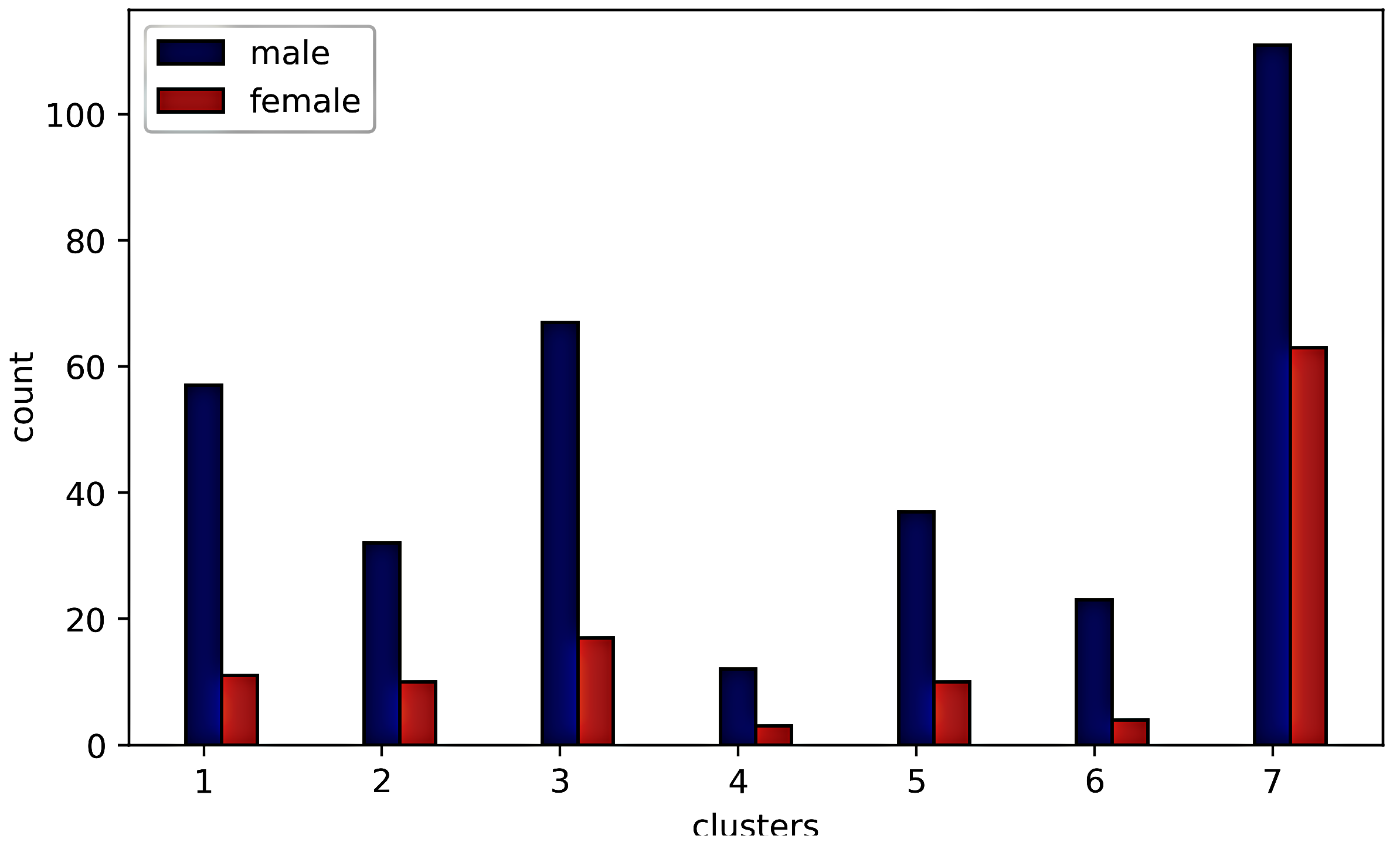}  
  \caption{Hierarchical clustering}
  \label{fig:sub-first}
\end{subfigure}
\begin{subfigure}{.5\textwidth}
  \centering
  \includegraphics[width=.95\linewidth]{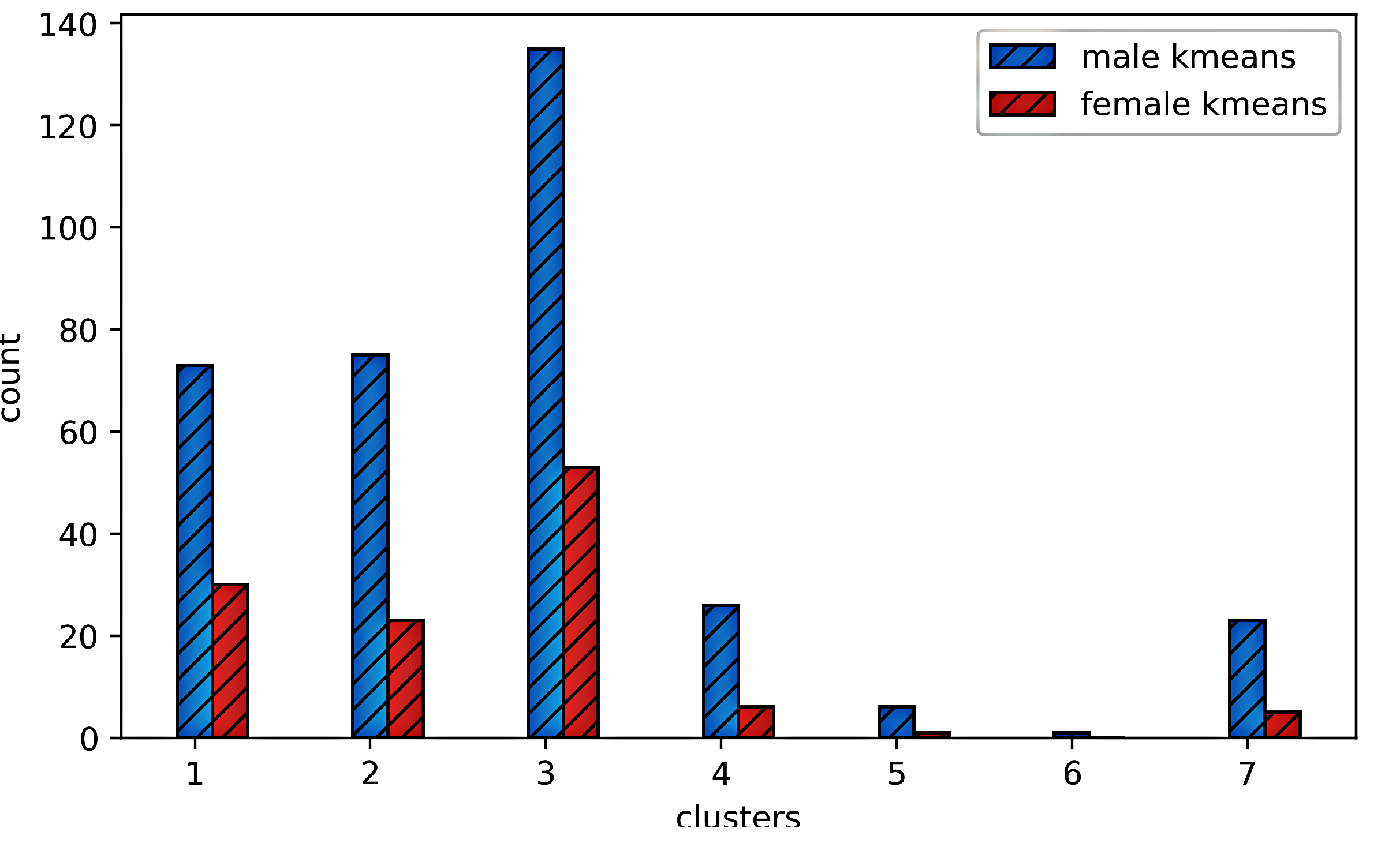}  
  \caption{Clustering with k means}
  \label{fig:sub-second}
\end{subfigure}
\caption{Results of clustering}
\label{fig:clustering}
\end{figure}

For further analysis to find out if there is an implicit bias between the traits of male and female characters, we have clustered characters together into seven smaller groups. The clusters are formed based on the similarity between characters, and we can safely say that the groups represent the type or nature of the characters in general. Therefore, characters with similar traits will be grouped and of somewhat the same category. The overall ratio of male and female characters in our database is 3:1. And if there were no bias in how the characters are written, the proportion of male and female characters would be similar in the clusters. However, we employed two different techniques to cluster them, and in both of them, there was an uneven ratio of genders in the smaller groups, as shown in Figure \ref{fig:clustering}. Some groups even have a balance of around 6:1, which implies high bias in writing those kinds of characters.

\section{Conclusion}
We have analysed gender representation in movies and how gender distribution changed over time. Through sentiment analysis, we have concluded that overall, male and female characters are shown to be somewhat equally positive and negative. But, through detailed research conducted with emotion detection, we have found that different kinds of positive emotions are associated with different genders. Those emotions reflect the stereotypes that exist in our society. For example, men are shown to be more dominant and envious than women, and women are shown to be more optimistic and joyful.

Compared to previous works, we rely not only on word vectors but create new embeddings by analysing the emotions associated with these words. We also look at how stereotypes are presented via emotions, making the results more intuitive. Furthermore, we highlighted the hidden bias by using clustering and other ML techniques.

\bibliography{sn-bibliography.bib}

\end{document}